\documentclass[10pt,twocolumn,letterpaper]{article}

\usepackage{cvpr}              

\usepackage{graphicx}
\usepackage{amsmath}
\usepackage{amssymb}
\usepackage{booktabs}
\usepackage{algorithm}  
\usepackage{algorithmicx}  
\usepackage{algpseudocode}  
\usepackage{multirow}
\usepackage{enumerate}
\usepackage{xcolor}
\usepackage{comment}
\usepackage[pagebackref,breaklinks,colorlinks]{hyperref}
\usepackage{colortbl}

\usepackage{ulem} 
\usepackage{adjustbox}
\usepackage[capitalize]{cleveref}
\crefname{section}{Sec.}{Secs.}
\Crefname{section}{Section}{Sections}
\Crefname{table}{Table}{Tables}
\crefname{table}{Tab.}{Tabs.}

\usepackage{enumitem}

\def\confName{CVPR}
\def\confYear{2024}

\begin{document}


\title{VLPose: Bridging the Domain Gap in Pose Estimation \\with Language-Vision Tuning}

\author{Jingyao Li$^{1}$ \quad\quad Pengguang Chen$^{2}$\quad\quad Xuan Ju$^{1}$\quad\quad Hong Xu$^{1}$ \quad\quad Jiaya Jia$^{1,2}$  \\[0.2cm]
 The Chinese University of Hong Kong$^{1}$\quad SmartMore$^{2}$\\
 jingyao.li@link.cuhk.edu.hk \quad  leojia@cse.cuhk.edu.hk
}
\maketitle

\begin{abstract}
Thanks to advances in deep learning techniques, Human Pose Estimation (HPE) has achieved significant progress in natural scenarios.
However, these models perform poorly in artificial scenarios such as painting and sculpture due to the domain gap, constraining the development of virtual reality and augmented reality. 
With the growth of model size, retraining the whole model on both natural and artificial data is computationally expensive and inefficient. 
Our research aims to bridge the domain gap between natural and artificial scenarios with efficient tuning strategies. 
Leveraging the potential of language models, we enhance the adaptability of traditional pose estimation models across diverse scenarios with a novel framework called VLPose. 
VLPose leverages the synergy between language and vision to extend the generalization and robustness of pose estimation models beyond the traditional domains. 
Our approach has demonstrated improvements of 2.26\% and 3.74\% on HumanArt and MSCOCO, respectively, compared to state-of-the-art tuning strategies.

\end{abstract}

\section{Introduction}
\label{sec:intro}

\begin{figure}[t]
  \centering
    \includegraphics[width=0.99\linewidth, trim=0 0 0 0, clip]{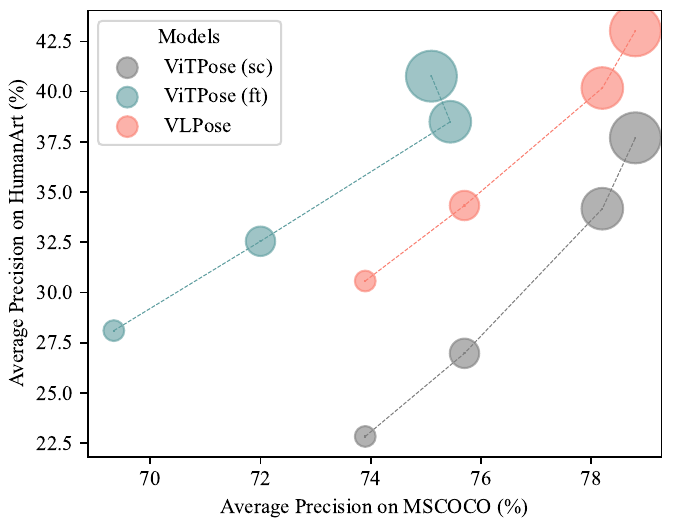}
    \caption{The comparison between our proposed VLPose, pretrained ViTPose~\cite{vitpose}, and finetuned ViTPose. \textit{ft} denotes the fine-tuned ViTPose model on HumanArt, and \textit{sc} represents the scratch ViTPose model pre-trained on MSCOCO. The size of the circle indicates the corresponding size of the model.}
  \label{fig:impress}
  \vspace{-0.3cm}
\end{figure}

Human Pose Estimation (HPE)~\cite{liu2022recent} has witnessed remarkable advancements in recent years. 
However, most of these endeavors target humans in natural settings~\cite{ionescu2013human3, li2019crowdpose, mscoco, narasimhaswamy2022whose}, and lack the generalization ability to other scenarios, particularly artificial scenarios such as painting and sculpture. 
For example, as shown in \cref{fig:impress}, current SOTA models pre-trained on natural scenes perform poorly on artificial scenes, and when fine-tuned on artificial scenes, they tend to experience a drop in their performance on natural scenarios.

The fundamental reason lies in the domain gap between natural and artificial scenes, as well as among various types of artificial scenes (animation, watercolor painting, shadow puppetry, sketching, etc.). The unsatisfying performance in artificial human scenes imposes substantial constraints on the development of various computer graphics-related tasks, including anime character image generation~\cite{chen2022improving, yang2022vtoonify, zheng2020learning}, character rendering~\cite{lin2022collaborative}, and character motion retargeting~\cite{aberman2020skeleton, mourot2022survey, yang2020transmomo}. With the growing interest and application of technologies such as virtual reality, augmented reality, and the metaverse, the domain gap in HPE has become more pronounced and requires immediate attention. In our research, our primary objective is to bridge the significant domain gap in natural and diverse artificial scenarios. 

\begin{figure*}[t]
  \centering
    \includegraphics[height=0.28\linewidth,width=0.99\linewidth, trim=95 40 115 300, clip]{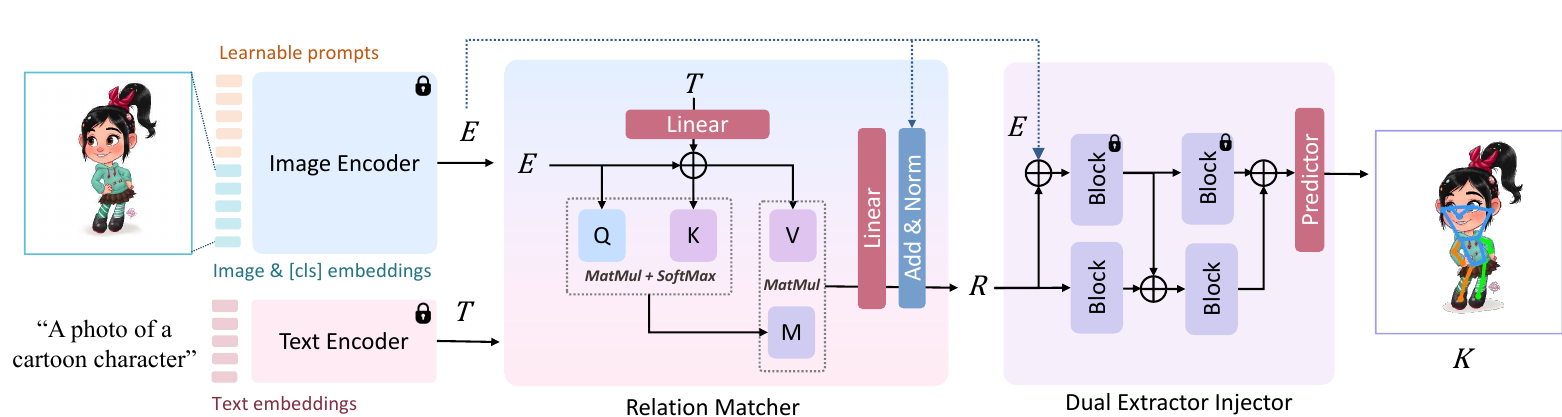}
    \caption{In our framework, a text encoder is utilized to encode domain-specific information. Then, a vision-language relation matcher captures the inter-relationship between images and text and feeds this information into the dual extractor injector for pose estimation. Domain-specific information is integrated into the pose estimator to minimize performance discrepancies stemming from different domains.}
  \label{fig:framework}
  \vspace{-0.3cm}
\end{figure*}

In recent times, the remarkable capabilities of language models~\cite{gpt3, gpt4, llava, blip2, minigpt4, lisa} in the domain of computer vision have gained widespread acknowledgment. A naive approach to bridging natural and artificial scenarios is to introduce a text encoder to encode domain-specific information. However, we observed that the straightforward concatenation of image and text features yielded limited improvements, as depicted in \cref{sec:ab_matcher}. Therefore, we introduced a \textit{Vision-Language Relation Matcher}, which effectively captures the inter-relationship between images and domain-specific text tokens, enhancing the model's overall performance. To leverage these image-text relations, we designed a novel vision-language decoder named the \textit{Dual Extractor-Injector Decoder} for pose estimation. The well-crafted decoder introduces an elevated level of interaction between vision and language, resulting in more effective and robust performance, as elaborated in \cref{sec:decoder}. 

To sum up, our vision-language pose estimation framework, named \textit{VLPose}, is depicted in \cref{fig:framework}. By incorporating domain-specific knowledge, our framework effectively mitigates performance discrepancies arising from the inherent diversity among different domains. As illustrated in \cref{fig:impress}, our proposed VLPose significantly improves the model's generalization ability across various artificial scenes without compromising its performance on natural scenes. Our contributions are threefold:

\begin{enumerate}

\item We propose a novel framework, VLPose, designed to address the domain gap between human-centric natural and artistic domains. Our method exhibits respective improvements of 2.26\% and 3.74\% compared to the current state-of-the-art on HumanArt and MSCOCO.

\item Our introduced vision-language relation matcher effectively captures and models the intricate interplay between images and domain-specific textual tokens, enhancing the model's overall performance.

\item Our proposed dual extractor-injector decoder integrates the vision-language relationship into the core of our pose estimator, further enhancing its performance and adaptability across diverse scenarios.
\end{enumerate}


\section{Related Work}
\label{sec:related}
\subsection{Pose Estimation Datasets}
In the field of human-centric computer vision datasets, there exist two main types: datasets focused on natural scenes and others dedicated to artificial scenarios. The fundamental tasks in human-centric recognition primarily involve human detection and pose estimation. Most established datasets~\cite{posetrack, 2d, pascal, crowdpose, lin2020human, narasimhaswamy2022whose, wu2019large, pose2seg}, meticulously annotate humans in natural settings using bounding boxes and keypoint annotations. Among these datasets, MSCOCO~\cite{mscoco} has gained preeminence, owing to its comprehensive coverage of diverse poses and intricate natural scenes. In contrast, datasets that revolve around artificial scenes are relatively sparse~\cite{sketch2pose, classarch, peopleart, humanart}. For instance, Sketch2Pose~\cite{sketch2pose} is designed for scenarios involving sketches, while ClassArch~\cite{classarch} exclusively features ancient vase paintings. People-Art~\cite{peopleart}, a dataset that encompasses both natural and artificial imagery, categorizes artificial scenes by directly incorporating artistic painting styles from wiki-art. Among these datasets, HumanArt~\cite{humanart} stands out as a representative dataset due to its rich scenario diversity, high-quality imagery, and versatile annotations.

Despite existing deep learning models have consistently demonstrated exceptional performance across a wide spectrum of natural scenes downstream tasks~\cite{jin2020whole, li2021human, moon2022accurate, vitpose}, there presents a challenge when these models are applied to artificial scenarios, where scene characteristics and contextual factors diverge significantly~\cite{humanart}. Through our research endeavors, we strive to leverage the capabilities of language models to overcome the limitations that previously constrained these models to specific domains.

\subsection{Pose Estimation Methods}
The field of pose estimation has witnessed rapid advancements, transitioning from Convolutional Neural Networks (CNNs)~\cite{wu2017ai} to Vision Transformer (ViT) networks. In the early stages, transformers were often considered as improved decoders~\cite{li2021pose, tokenpose, transpose}. For instance, TransPose~\cite{transpose} processed features extracted by CNNs directly to capture global relationships, while TokenPose~\cite{tokenpose} introduced extra tokens to estimate occluded keypoint locations and model relationships between different key points through token-based representations. HRFormer~\cite{hrformer} emerged as a solution to eliminate the need for CNNs in feature extraction, relying on transformers to directly extract high-resolution features. Additionally, ViTPose~\cite{vitpose} delved into the potential of plain vision transformers in pose estimation tasks, introducing a simple yet effective baseline model based on these transformers.

In our work, to the best of our knowledge, we represent the first concerted effort to address the disparities between natural and artificial scenes within the context of human-centric recognition. Furthermore, it stands as the pioneering initiative to integrate language models with pose estimation models. This unique amalgamation has propelled our method to achieve exceptional cross-domain performance.

\section{Method}
\label{sec:method}

In our research, we present a framework known as VLPose, which is illustrated in \cref{fig:framework}. We introduce our image encoder in \cref{sec:image_encoder} and text encoder in \cref{sec:text_encoder}. Subsequently, a vision-language relationship matcher captures the intricate connections existing between images and text in \cref{sec:matcher}. Then this knowledge is funneled into our vision-language decoder in \cref{sec:decoder}. 

\subsection{Image Encoder}
\label{sec:image_encoder}

Within our framework, We denote the features of the image encoder as $E \in \mathbb{R}^{P\times C}$, where $P$ is the number of image patches, and $C$ is its channel dimension. We employ a visual prompt tuning strategy~\cite{vpt}, which entails freezing the original model weights and incorporating an image prompt into the architecture, as shown in \cref{fig:framework}. One notable advantage of this approach is its reversibility, as we can seamlessly revert to the original model's performance by merely removing the added visual prompt. It empowers us to strike a balance between the model's proficiency in real-world scenarios and virtual environments. By toggling the presence of the visual prompt, we can effortlessly tailor the model's behavior to suit specific tasks and domains, without sacrificing its fundamental capabilities.

\subsection{Text Encoder}
\label{sec:text_encoder}

We employ a text encoder to encode domain-specific text prompts. These encoded features are represented as $T \in \mathbb{R}^{L\times D}$, where $L$ signifies the length of the text tokens, and $D$ represents the channel dimension. We have meticulously designed specific prompts for each category. Leveraging pre-trained language models, which possess a deep understanding of these prompts, provides our model with valuable features. This enables our model to achieve significant performance enhancements across a wide range of scenarios, thus showcasing its versatility and effectiveness in diverse contexts. In \cref{sec:ab_encoder}, we conduct an experimental exploration of the impact of domain prompts on our model's performance. 

\begin{figure*}[t]
  \centering
  \begin{subfigure}{0.24\linewidth}
    \includegraphics[width=0.99\linewidth, trim=40 240 715 165, clip]{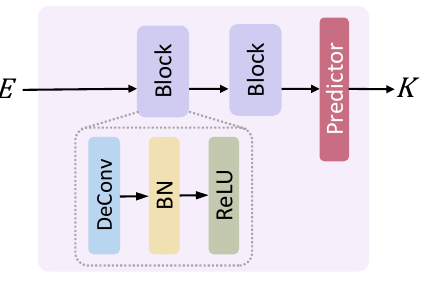}
    \caption{Baseline}
    \label{fig:decoder_baseline}
\end{subfigure}
\begin{subfigure}{0.24\linewidth}
    \includegraphics[width=0.99\linewidth, trim=255 240 500 165, clip]{figures/4_decoder.pdf}
    \caption{Injector}
    \label{fig:decoder_in}
\end{subfigure}
\begin{subfigure}{0.24\linewidth}
    \includegraphics[width=0.99\linewidth, trim=468 240 287 165, clip]{figures/4_decoder.pdf}
    \caption{Extractor-Injector}
    \label{fig:decoder_ex_in}
\end{subfigure}
\begin{subfigure}{0.24\linewidth}
    \includegraphics[width=0.99\linewidth, trim=692 240 63 165, clip]{figures/4_decoder.pdf}
    \caption{Dual-Extractor-Injector}
    \label{fig:decoder_2_ex_in}
\end{subfigure}
\caption{A spectrum of decoder architectures: (a) The baseline decoder features two deconvolution blocks, upsampling operations, and a $1\times1$ convolution layer to generate keypoints' heatmaps. (b) The injector decoder combines image feature $E$ and image-text relation $R$, feeding them into the baseline structure. (c) The extractor-injector decoder has two branches: main and auxiliary branch. The auxiliary branch extracts features and injects relationship knowledge to improve pose estimation. (d) The dual extractor-injector decoder strengthens the interaction between main and auxiliary branches. }
\label{fig:decoder}

\end{figure*}

\subsection{Relation Matcher}
\label{sec:matcher}
To establish a robust connection between vision and language features, we introduce a straightforward architecture called \textit{Relation Matcher}, as illustrated in \cref{fig:framework}. The relation matcher encompasses a multi-head attention mechanism, coupled with two linear layers and a normalization layer. The text features, encoded by the text encoder, undergo an initial transformation via a linear layer to align their dimensions with those of the image features:
\begin{equation}
    T' = \phi_T(T) \in \mathbb{R}^{L\times C},
\end{equation}
where $\phi_T$ is the linear layer of $T$; $L$ is the length of the text tokens and $C$ is the channel dimension of the images features. Subsequently, the transformed text features $T'$ are concatenated with the original image features $E$.
\begin{equation}
    \hat{T} = {\rm concat}[E, T'].
\end{equation}
The composite input $\hat{T}$ is then fed into the multi-head attention mechanism, whose query $Q$, key $K$ and value $V$ are:
\begin{equation}
\begin{aligned}
Q &= E \in \mathbb{R}^{P\times C}, \\
K &= \hat{T} \in \mathbb{R}^{L\times C}, \\
V &= \hat{T} \in \mathbb{R}^{L\times C}.
\end{aligned}
\end{equation}
Consequently, the relation matcher produces the image-text relation $R$ as:
\begin{equation}
\begin{aligned}
R' &= {\rm softmax}(\frac{QK^T}{\sqrt{C}})V, \\
R &= {\rm norm}(\phi_R(R') + Q),
\end{aligned}
\end{equation}
where $\phi_R$ is the linear layer of $R$; ${\rm norm}$ and ${\rm softmax}$ refers to the normalization layer and softmax layer. $R'$ represents the output of the attention block. Our relation matcher facilitates the learning of intricate vision-language relationships, resulting in improved performance across various tasks and scenarios. In \cref{sec:ab_matcher}, we conduct comprehensive ablation experiments to examine and substantiate the effectiveness of our proposed structure. 

\subsection{Vision-Language Decoder}
\label{sec:decoder}
By feeding the vision and language features into the relation matcher, we obtain the image-text relationship matrix \(R \in \mathbb{R}^{P \times C}\), which is then input into the pose decoder. In this section, we will outline the structure of the pose decoder, encompassing both the traditional pose estimator and our proposed vision-language decoders. The architectural details are depicted in \cref{fig:decoder}.

\paragraph{Baseline.} The baseline decoder is comprised of two deconvolution blocks, each containing a deconvolutional layer, followed by batch normalization~\cite{bn}, and activation using the Rectified Linear Unit (ReLU)~\cite{relu}.   Following established conventions from prior studies~\cite{zhang2020empowering, baseline}, each block executes an upsampling operation, effectively doubling the size of the feature maps. Subsequently, a convolution layer with a $1\times1$ kernel size is employed to generate the localization heatmaps for the keypoints. The specifics of the architecture are visually represented in \cref{fig:decoder_baseline}. Mathematically, this process can be expressed as 
\begin{equation}\label{equ:decoder_baseline}
K_{b} = p(f_m(f_m(E))), 
\end{equation}
where $K_{b} \in \mathbb{R}^{\frac{H}{4} \times\frac{W}{4} \times N_k}$ represents the estimated heatmaps, with one heatmap produced for each keypoint, and $N_k$ denoting the total number of keypoints to be estimated. $E$ denotes the image feature. $p$ refers to the predictor, which consists of a $1\times1$ convolution layer. $f_m$ represents a decoder block, with the subscript $m$ serves to distinguish it as the main branch, setting it apart from the auxiliary branch, which will be discussed later.

\paragraph{Injector Decoder.} An injector decoder involves a simple method to inject image-text relationship knowledge into the traditional decoder. It combines the image feature $E$ with the image-text relation $R$ and subsequently feeds the combined information into the structure identical to the baseline. \cref{fig:decoder_in} illustrates the design. The procedure can be formally represented as
\begin{equation}\label{equ:decoder_in}
K_{i} = p(f_m(f_m(E+R))), 
\end{equation}
where $p$ and $f_m$ denote the predictor and the decoder block.

\paragraph{Extractor-Injector Decoder.} Drawing inspiration from the successful developments within the field of dense prediction tasks~\cite{vitadapter}, we have devised an approach tailored specifically for our vision-language decoder. The decoder is primarily composed of two branches, each consisting of 2 decoder blocks. The main branch is labeled as $f_m$, while the auxiliary branch is denoted as $f_a$. The role of the auxiliary branch is to extract image features and inject relationship knowledge into the main branch, aiding in pose estimation through an extractor-injector mechanism. A visual representation of the architecture is in \cref{fig:decoder_ex_in}. The output of the main branch is as follows:
\begin{equation}\label{equ:decoder_ex_in}
K_{ei} = p(f_m(f_m(E)) + f_a(f_a(E+R))), 
\end{equation}

Since the main branch shares the same structure as the traditional pose decoder, it's evident that when $R$ equals 0 and the auxiliary branch weights are all zeros ($f_a$ weights), the VLPose estimator reverts to the baseline Pose decoder as \cref{equ:decoder_baseline}. Therefore, when we freeze the main branch and fine-tune the auxiliary branch, we can still preserve the capability of the original structure.

\begin{table}[t]
\centering
\setlength{\tabcolsep}{1mm}
\small
\begin{tabular}{c|ccccc}
\toprule
\multirow{2}{*}{Model} & batch & learning & weight & layer-wise & drop path \\
 & Size & rate & decay & decay & rate \\
\midrule
VLPose-S & 512 & $5\times10^{-4}$ & 0.1 & 0.75 & 0.30 \\
VLPose-B & 512 & $5\times10^{-4}$ & 0.1 & 0.75 & 0.30 \\
VLPose-L & 512 & $5\times10^{-4}$ & 0.1 & 0.80 & 0.50 \\
VLPose-H & 512 & $5\times10^{-4}$ & 0.1 & 0.80 & 0.55 \\
\bottomrule
\end{tabular}
\caption{Hyper-parameters for training VLPose.}
\label{tab:cfg}
\end{table}

\begin{table*}[t]
\centering
\begin{subtable}[t!]{0.48\linewidth}
\centering
\setlength{\tabcolsep}{2.3mm}
\small
\vspace{-2mm}
\begin{tabular}{c|ccccc}
\toprule
$K=V$ & AP & AP50 & AP75 & AR & AR50 \\
\midrule
w/o text & 36.45 & 52.60 & 38.56 & 43.12 & 59.28 \\
w/o matcher & 36.54 & 52.98 & 38.48 & 43.09 & 59.34 \\
\rowcolor{gray!20}w matcher & \textbf{37.16} & \textbf{53.14} & \textbf{39.43} & \textbf{43.76} & \textbf{59.90} \\
\bottomrule
\end{tabular}
\caption{Relation Matcher.} 
\label{tab:wo_matcher}
\end{subtable}
\hspace{2mm}
\vspace{2mm}
\begin{subtable}[t!]{0.48\linewidth}
\vspace{-2mm}
\centering
\setlength{\tabcolsep}{2.8mm}
\small
\begin{tabular}{c|ccccc}
\toprule
Model & AP & AP50 & AP75 & AR & AR50 \\
\midrule
None & 36.45 & 52.60 & 38.56 & 43.12 & 59.28 \\
albef~\cite{albef}& 36.78 & 52.63 & 38.78 & 43.49 & 59.40 \\
\rowcolor{gray!20}blip~\cite{blip}& \textbf{37.16} & \textbf{53.14} & \textbf{39.43} & \textbf{43.76} & \textbf{59.90} \\
\bottomrule
\end{tabular}
\caption{Text encoder.}
\label{tab:encoder}
\end{subtable}

\begin{subtable}[t!]{0.48\linewidth}
\centering
\setlength{\tabcolsep}{2.7mm}
\small
\begin{tabular}{c|ccccc}
\toprule
$K=V$ & AP & AP50 & AP75 & AR & AR50 \\
\midrule
None & 36.45 & 52.60 & 38.56 & 43.12 & 59.28 \\
$T$ & 36.58 & 52.98 & 38.68 & 43.23 & 59.70 \\
$[E, E\cdot T]$ & 36.74 & 53.02 & 38.74 & 43.49 & \textbf{59.91} \\
\rowcolor{gray!20}$[E, T]$ & \textbf{37.16} & \textbf{53.14} & \textbf{39.43} & \textbf{43.76} & 59.90 \\
\bottomrule
\end{tabular}
\caption{Multi-head attention input configurations in the relation matcher.}
\label{tab:matcher}
\end{subtable}
\hspace{2mm}
\begin{subtable}[t!]{0.48\linewidth}
\small
\centering
\setlength{\tabcolsep}{2.5mm}
\begin{tabular}{c|ccccc}
\toprule
Prompt & AP & AP50 & AP75 & AR & AR50 \\
\midrule
None & 36.45 & 52.60 & 38.56 & 43.12 & 59.28 \\
Random & 36.74 & 52.77 & 38.90 & 43.30 & 59.36 \\
Fixed prompt & 36.76 & 52.78 & 39.20 & 43.45 & 59.52 \\
\rowcolor{gray!20}Style prompt & \textbf{37.16} & \textbf{53.14} & \textbf{39.43} & \textbf{43.76} & \textbf{59.90} \\
\bottomrule
\end{tabular}
\caption{Text prompt.}
\setlength{\tabcolsep}{2.5mm}
\label{tab:prompt}
\end{subtable}

\vspace{2mm}
\begin{subtable}[t!]{0.48\linewidth}
\centering
\setlength{\tabcolsep}{3mm}
\small
\begin{tabular}{c|ccccc}
\toprule
Model & AP & AP50 & AP75 & AR & AR50 \\
\midrule
None & 36.45 & 52.60 & 38.56 & 43.12 & 59.28 \\
in & 36.93 & 52.86 & 39.34 & 43.48 & \uline{59.55} \\
ex-in & \uline{37.13} & \uline{53.08} & \textbf{39.47} & \uline{43.63} & 59.48 \\
\rowcolor{gray!20} 2-ex-in & \textbf{37.16} & \textbf{53.14} & \uline{39.43} & \textbf{43.76} & \textbf{59.90} \\
\bottomrule
\end{tabular}
\caption{Vision-Language Decoder.}
\label{tab:decoder}
\end{subtable}
\hspace{2mm}
\begin{subtable}[t!]{0.48\linewidth}
\centering
\setlength{\tabcolsep}{4.5mm}
\small
\begin{tabular}{c|cccccc}
\toprule
Tokens & Small & Base & Large & Huge \\
\midrule
5 & 30.05 & 34.05 & 40.12 & 42.83 \\
10 & 30.16 & 34.31 & \cellcolor{gray!20}\textbf{40.18} & \cellcolor{gray!20}\textbf{43.03} \\
20 & 30.31 & \cellcolor{gray!20}\textbf{34.33} & 40.12 & 42.78 \\
50 & \cellcolor{gray!20}\textbf{30.57} & 34.17 & 39.82 & - \\
\bottomrule
\end{tabular}
\caption{Number of learnable visual tokens.}
\label{tab:num_tokens}
\end{subtable}

\caption{Ablation experiments. (a) \textbf{Relation matcher architecture.} \textit{w/o text} is the naive baseline without integrating text tokens. \textit{w/o matcher} simply concatenates text and image tokens. (b) \textbf{Text encoder}, including albef~\cite{albef} and blip~\cite{blip}. (c) \textbf{Multi-head input configuration} in the relation matcher, including text feature $T$, image feature $E$, the concatenation of both features $[E, T]$ and the concatenation of image feature and cosine similarity of the features $[E, E\cdot T]$.  (d) \textbf{Text prompts}, including random prompts, fixed prompts, and style prompts. (e) \textbf{Pose decoder structures}, encompassing the baseline decoder (\textit{None}), injector decoder (\textit{in}), extractor-injector decoder (\textit{ex-in}), and dual extractor-injector decoder (\textit{2-ex-in}). (f) \textbf{Number of learnable visual tokens} for difference model sizes. Bold and underlined text signify the best and second-best results. Gray background indicates our selection.}
\end{table*}

\paragraph{Dual Extractor-Injector Decoder.} Furthermore, we have advanced our approach by introducing a decoder that incorporates a dual extractor-injector structure, which is graphically presented in \cref{fig:decoder_2_ex_in}. In this configuration:
\begin{equation}
\begin{aligned}
    K'_{2ei} &= f_m(E+R), \\
    K_{2ei} &= p(f_m(K'_{2ei}) + f_a(K'_{2ei}+f_a(R))), 
\end{aligned}
\end{equation}
where $p$ denotes the predictor. $f_m$ and $f_a$ represent the main and auxiliary branch. 

The dual extractor-injector decoder combines the advantages of the extractor-injector decoder, offering the flexibility to restore pre-trained weights by simply setting $R$ to zero and omitting the auxiliary branch. This approach preserves the performance of pre-training while allowing for model fine-tuning. Moreover, the dual extractor-injector decoder introduces an elevated level of synergy and cooperation between the main branch and the auxiliary branch. This heightened interaction serves to augment the model's overall capacity, leading to more effective and robust performance across various tasks and domains. 

In \cref{sec:ab_decoder}, we execute a series of ablation experiments to validate the advantages conferred by our extractor-injector structures when contrasted with conventional architectures. In the appendix, we further delve into additional strategies for fusing multi-modal knowledge.

\section{Experiments}
\label{sec:exp}
VLPose adheres to the conventional top-down approach for human pose estimation, where a detector is first employed to identify individuals within an image, and then VLPose is utilized to estimate the keypoint locations for these detected individuals.

\paragraph{Configuration.} We follow the default training configuration of previous works~\cite{vitpose, pvt, swin, pvtv2}, which includes using a 256$\times$192 input resolution and the AdamW~\cite{AdamW} optimizer with a learning rate of $5\times10^{-4}$. Post-processing is performed using Udp~\cite{udp}. The models undergo training for 210 epochs, with a learning rate reduction by a factor of 10 at the 170th and 200th epochs. The layer-wise learning rate decay~\cite{xlnet} and the stochastic drop path ratio for each model are presented in~\cref{tab:cfg}.

\subsection{Ablation}
\label{sec:ab}
In this section, we conduct an extensive series of ablation experiments to substantiate the effectiveness of each module within our approach, including the relation matcher (\cref{sec:ab_matcher}), vision-language decoder (\cref{sec:ab_decoder}), text encoder (\cref{sec:ab_encoder}) and finetuning method (\cref{sec:finetune}). We utilize the ViT-S~\cite{vit} as the backbone and conducted training for each model from scratch on HumanArt~\cite{humanart} for ablation experiments of our proposed method.

\subsubsection{Relation Matcher}
\label{sec:ab_matcher}
Firstly, we validate the effectiveness of our proposed relation matcher. We compare with a straightforward method that directly feeds the concatenation of vision and language features to the decoder without going through the relation matcher. As shown in \cref{tab:wo_matcher}, it does not effectively utilize the concatenated features, resulting in a marginal improvement of 0.09\% only. However, incorporating our relation matcher introduces a notable increase of 0.71\%.

Then, we systematically investigated various input configurations for the multi-head attention's key and value components. The results, as presented in \cref{tab:matcher}, clearly demonstrate that the concatenated representation of both features, denoted as $[E, T]$, yields the most promising performance across multiple evaluation metrics, including Average Precision (AP) and Average Recall (AR) at various thresholds. It demonstrates that the relation matcher is able to take advantage of the straightforward image and text concatenation to significantly enhance the overall performance of the pose estimation.

\begin{table}[t]
\centering
\setlength{\tabcolsep}{2mm}
\small
\begin{subtable}[t]{\linewidth}
\begin{tabular}{c|c|cccc}
\toprule
\multirow{2}{*}{Model} & \multirow{2}{*}{Finetune} & \multicolumn{4}{c}{Size} \\
 &  & Small & Base & Large & Huge \\
 \midrule
\multirow{6}{*}{ViTPose \cite{vitpose}} & - & 22.83 & 26.97 & 34.17 & 37.7 \\
 & 5 tokens & 27.71 & 32.38 & 38.37 & 40.81 \\
 & 10 tokens & 27.88 & 32.49 & 38.50 & 40.77 \\
 & 20 tokens & 27.96 & 32.55 & 38.65 & 40.89 \\
 & 50 tokens & 28.10 & 32.54 & 38.61 & - \\
 & last layer & \uline{29.82} & \uline{33.93} & \uline{39.46} & \uline{41.32} \\
  \midrule
\rowcolor{gray!20}\multicolumn{2}{c|}{VLPose} & \textbf{30.57} & \textbf{34.33} & \textbf{40.18} & \textbf{43.03} \\
\bottomrule
\end{tabular}
\caption{Average Precision (AP) of finetuning methods.} 
\label{tab:finetune}
\end{subtable}
\vspace{2mm}

\begin{subtable}[t]{\linewidth}
\centering
\setlength{\tabcolsep}{1.5mm}
\small
\begin{tabular}{c|c|cccc}
\toprule
\multirow{2}{*}{Model} & \multirow{2}{*}{finetune} & \multicolumn{4}{c}{Size} \\
 &  & Small & Base & Large & Huge \\
 \midrule
\multirow{2}{*}{ViTPose \cite{vitpose}} & visual prompt & 2.65 & 4.22 & 5.26 & 6.30 \\
 & last layer & 4.40 & 11.29 & 17.85 & 25.98 \\
   \midrule
\rowcolor{gray!20}VLPose & visual prompt & 3.74 & 6.75 & 10.48 & 14.05 \\
\bottomrule
\end{tabular}
\caption{Trainable parameters (MB) of finetuning methods.} 
\label{tab:param}
\end{subtable}
\caption{Performance and parameters of different finetuning methods including visual prompt tuning and last-layer tuning.} 
\end{table}

\begin{table*}[t]
\centering
\setlength{\tabcolsep}{4.3mm}
\small
\begin{tabular}{c|cc|cc|cccc}
\toprule
 & \multicolumn{2}{c|}{HumanArt} & \multicolumn{2}{c|}{MSCOCO} & \multicolumn{2}{c}{Average} \\
\multirow{-2}{*}{Method} & AP & AP$\ddagger $ & AP & AP$\ddagger $ & AP & AP$\ddagger $ \\
\midrule
ViTPose-S (sc) & 22.83 & 50.71 & 73.92 & 75.94 & 48.38 & 63.33 \\
ViTPose-S (pt) & 28.1 & 62.61 & 69.34 & 71.22 & 48.72 & 66.91 \\
\rowcolor{gray!20}VLPose-S & \textbf{30.57} \scriptsize{\color{teal}(+2.47)} & \textbf{67.13} \scriptsize{\color{teal}(+4.52)} & \textbf{73.92} \scriptsize{\color{teal}(+4.58)} & \textbf{75.94} \scriptsize{\color{teal}(+4.72)} & \textbf{52.25} \scriptsize{\color{teal}(+3.53)} & \textbf{71.53} \scriptsize{\color{teal}(+4.62)} \\
ViTPose-B (sc) & 26.96 & 55.45 & 75.75 & 77.94 & 51.36 & 66.69 \\
ViTPose-B (pt) & 32.55 & 66.79 & 72.0 & 74.14 & 52.27 & 70.47 \\
\rowcolor{gray!20}VLPose-B & \textbf{34.33} \scriptsize{\color{teal}(+1.78)} & \textbf{70.36} \scriptsize{\color{teal}(+3.57)} & \textbf{75.75} \scriptsize{\color{teal}(+3.75)} & \textbf{77.94} \scriptsize{\color{teal}(+3.80)} & \textbf{55.04} \scriptsize{\color{teal}(+2.77)} & \textbf{74.15} \scriptsize{\color{teal}(+3.69)} \\
ViTPose-L (sc) & 34.17 & 63.71 & 78.18 & 80.74 & 56.18 & 72.22 \\
ViTPose-L (pt) & 38.5 & 71.86 & 75.44 & 78.16 & 56.97 & 75.01 \\
\rowcolor{gray!20}VLPose-L & \textbf{40.18} \scriptsize{\color{teal}(+1.68)} & \textbf{74.95} \scriptsize{\color{teal}(+3.09)} & \textbf{78.18} \scriptsize{\color{teal}(+2.74)} & \textbf{80.74} \scriptsize{\color{teal}(+2.58)} & \textbf{59.18} \scriptsize{\color{teal}(+2.21)} & \textbf{77.84} \scriptsize{\color{teal}(+2.84)} \\
ViTPose-H (sc) & 37.7 & 66.46 & 78.84 & 81.4 & 58.27 & 73.93 \\
ViTPose-H (pt) & 40.77 & 73.55 & 75.1 & 77.33 & 57.94 & 75.44 \\
\rowcolor{gray!20}VLPose-H & \textbf{43.03} \scriptsize{\color{teal}(+2.26)} & \textbf{76.41} \scriptsize{\color{teal}(+2.86)} & \textbf{78.84} \scriptsize{\color{teal}(+3.74)} & \textbf{81.40} \scriptsize{\color{teal}(+4.07)} & \textbf{60.94} \scriptsize{\color{teal}(+3.00)} & \textbf{78.91} \scriptsize{\color{teal}(+3.47)} \\
\midrule
SwinPose-T (sc) & 21.98 & 49.86 & 72.44 & 74.03 & 47.21 & 61.95 \\
SwinPose-T (pt) & 26.01 & 59.03 & 70.06 & 71.38 & 48.04 & 65.20 \\
\rowcolor{gray!20}SwinVLPose-T & \textbf{27.81} \scriptsize{\color{teal}(+1.80)} & \textbf{62.53} \scriptsize{\color{teal}(+3.50)} & \textbf{72.44} \scriptsize{\color{teal}(+2.38)} & \textbf{74.03} \scriptsize{\color{teal}(+2.65)} & \textbf{50.12} \scriptsize{\color{teal}(+2.09)} & \textbf{68.28} \scriptsize{\color{teal}(+3.08)} \\
SwinPose-B (sc) & 26.88 & 50.96 & 73.74 & 75.74 & 50.31 & 63.35 \\
SwinPose-B (pt) & 31.01 & 61.38 & 72.55 & 74.13 & 51.78 & 67.75 \\
\rowcolor{gray!20}SwinVLPose-B & \textbf{31.92} \scriptsize{\color{teal}(+0.91)} & \textbf{62.41} \scriptsize{\color{teal}(+1.03)} & \textbf{73.74} \scriptsize{\color{teal}(+1.19)} & \textbf{75.74} \scriptsize{\color{teal}(+1.61)} & \textbf{52.83} \scriptsize{\color{teal}(+1.05)} & \textbf{69.07} \scriptsize{\color{teal}(+1.32)} \\
SwinPose-L (sc) & 28.37 & 53.22 & 74.32 & 76.3 & 51.34 & 64.76 \\
SwinPose-L (pt) & 33.21 & 62.36 & 73.56 & 75.01 & 53.39 & 68.69 \\
\rowcolor{gray!20}SwinVLPose-L & \textbf{34.17} \scriptsize{\color{teal}(+0.96)} & \textbf{62.98} \scriptsize{\color{teal}(+0.62)} & \textbf{74.32} \scriptsize{\color{teal}(+0.76)} & \textbf{76.30} \scriptsize{\color{teal}(+1.29)} & \textbf{54.24} \scriptsize{\color{teal}(+0.86)} & \textbf{69.64} \scriptsize{\color{teal}(+0.95)} \\
\midrule
PVT-S (sc) & 19.85 & 45.79 & 71.42 & 72.68 & 45.64 & 59.23 \\
PVT-S (pt) & 25.21 & 56.87 & 70.8 & 71.59 & 48.00 & 64.23 \\
\rowcolor{gray!20}PVLT-S & \textbf{26.04} \scriptsize{\color{teal}(+0.83)} & \textbf{57.98} \scriptsize{\color{teal}(+1.11)} & \textbf{71.42} \scriptsize{\color{teal}(+0.62)} & \textbf{72.68} \scriptsize{\color{teal}(+1.09)} & \textbf{48.73} \scriptsize{\color{teal}(+0.73)} & \textbf{65.33} \scriptsize{\color{teal}(+1.10)} \\
PVTv2-B (sc) & 24.71 & 52.91 & 73.65 & 75.68 & 49.18 & 64.30 \\
PVTv2-B (pt) & 31.09 & 60.03 & 72.28 & 74.31 & 51.69 & 67.17 \\
\rowcolor{gray!20}PVLT-B & \textbf{32.76} \scriptsize{\color{teal}(+1.67)} & \textbf{62.11} \scriptsize{\color{teal}(+2.08)} & \textbf{73.65} \scriptsize{\color{teal}(+1.37)} & \textbf{73.65} \scriptsize{\color{teal}(+-0.66)} & \textbf{53.20} \scriptsize{\color{teal}(+1.52)} & \textbf{67.88} \scriptsize{\color{teal}(+0.71)} \\
\bottomrule
\end{tabular}
\caption{Comparison of Average Precision (AP) among widely used human pose estimation models. With or without $\ddagger$ denotes the top-down pose estimation results utilizing ground truth and YOLO \cite{yolo} detected bounding boxes. Our experiments encompass evaluations on both MSCOCO and HumanArt datasets. \textit{sc}: scratch model pre-trained on MSCOCO. \textit{pt}: prompt finetuned model on HumanArt.}
\label{tab:result}
\end{table*}

\subsubsection{Vision-Language Decoder}
\label{sec:ab_decoder}
In \cref{sec:decoder}, we provided an in-depth overview of pose decoder structures: the baseline decoder, injector decoder, extractor-injector decoder, and dual extractor-injector decoder. To comprehensively evaluate their performance and understand their contributions, we conducted a series of ablation experiments. The results, as presented in \cref{tab:decoder}, revealed noteworthy insights into the effectiveness of these decoder structures. It is obvious that all three variants of the vision-language decoder configurations displayed improvements over the baseline decoder. This highlights the significance of introducing text features in bolstering the model's ability to perform across various domains and scenarios. 

Comparing the injector decoder and the extractor-injector decoder, we observe that the latter harnesses relations more effectively. This is attributed to the introduction of an auxiliary branch, which aids in extracting and incorporating relationship knowledge into the main branch. This design capitalizes on the synergy between vision features and textual relations, resulting in superior performance. Notably, the dual extractor-injector decoder achieves the highest level of performance. This accomplishment can be attributed to its heightened emphasis on interaction and collaboration between the main branch and the auxiliary branch. In the appendix, we explore various modes of interaction between the two branches, confirming the approach we have adopted to be the most beneficial for the model's utilization of cross-modal knowledge.

\subsubsection{Text Encoder}
\label{sec:ab_encoder}
To investigate the influence of text features on our results, we conducted ablation experiments employing different text encoders, as illustrated in \cref{tab:encoder}. All text feature representations yield performance improvements compared to the baseline, with BLIP~\cite{blip} achieving the most outstanding performance. Consequently, we integrate BLIP as the text encoder within our framework.

Furthermore, to explore the impact of various text prompts on our results, we carry out ablation experiments. Results are in \cref{tab:prompt}. Initially, we seek to demonstrate that the enhancements in performance attributed to text features were not solely a consequence of an increased number of model parameters. To substantiate this, we conduct ablation experiments involving random prompts while maintaining consistent settings, resulting in parameters identical to our framework. The outcomes clearly indicate that augmenting the number of parameters led to relatively limited improvements in performance. In contrast, the primary source of performance enhancement stemmed from the incorporation of specific text prompts. This advantage can be attributed to the nuanced understanding of semantics facilitated by meticulously pre-trained language models. Consequently, these models provided text features conducive to enhancing the overall architecture. Moreover, tailoring prompts to each semantic context further elevated our results, underscoring the significance of domain-specific prompts in achieving enhanced performance across different semantic environments, as detailed in the appendix.

\begin{figure*}[t]
  \centering
    \includegraphics[width=0.99\linewidth, trim=0 0 0 0, clip]{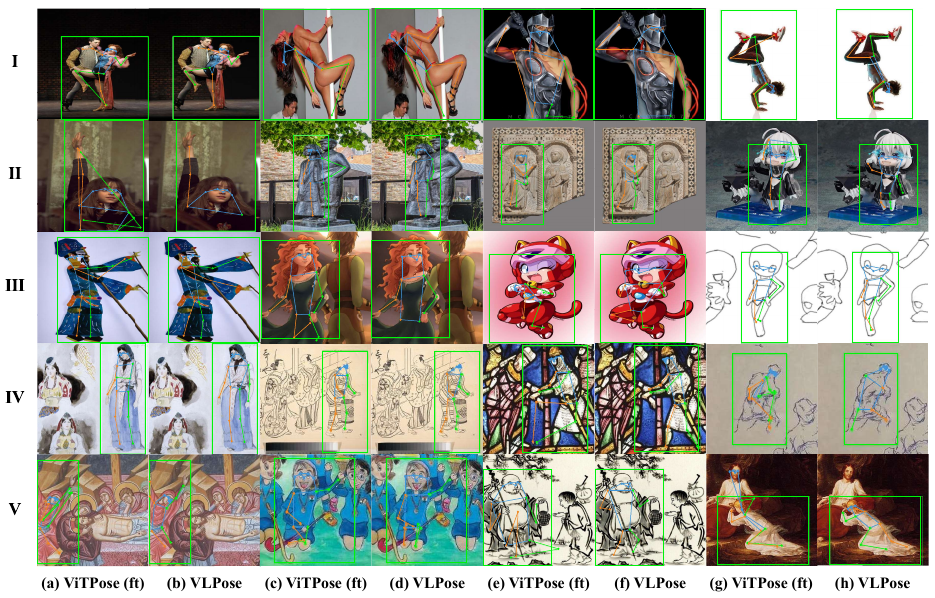}
    \caption{A visual comparison between our method and the current SOTA, ViTPose~\cite{vitpose}, on categories of HumanArt~\cite{humanart} involving real human bodies with large motion amplitudes in \uppercase\expandafter{\romannumeral1}(a)-\uppercase\expandafter{\romannumeral2}(a), 3D artistic human bodies in \uppercase\expandafter{\romannumeral2}(c)-\uppercase\expandafter{\romannumeral2}(g), and 2D artistic human bodies in \uppercase\expandafter{\romannumeral3}-\uppercase\expandafter{\romannumeral5}.}
  \label{fig:vis}
\vspace{-2mm}
\end{figure*}

\subsubsection{Finetuning Method}
\label{sec:finetune}
In order to preserve the model's generalization capability, we seek to avoid disrupting the original model weights. Thus, our approach involves finetuning through learnable visual prompt tuning. According to the experiments in \cref{tab:num_tokens}, we selected the optimal number of learnable image tokens for each size. Our results, as shown in \cref{tab:finetune}, demonstrate significant improvements across all sizes when compared to the current SOTA. Specifically, our method yields a 2.14\% performance boost on ViT-H when finetuning the same 50 tokens as the current SOTA model.

Furthermore, we perform finetuning on the final layer of the current SOTA model, introducing a larger number of parameters. This enhancement results in a more favorable outcome for competitive methods. Nevertheless, our method, while reducing the parameter count by 46\%, still achieves a 1.71\% performance boost for ViT-H. More parameter comparisons are presented in \cref{tab:param}.

\subsection{Results}
\label{sec:results}

In this section, models are initialized with pre-trained weights from MSCOCO~\cite{mscoco} and finetuned on HumanArt~\cite{humanart}. We freeze the original weights and finetune the visual prompts and auxiliary branches of the vision-language decoder. The comparisons were systematically applied to models of various sizes, and extensive evaluations were conducted on both the MSCOCO and HumanArt. 

Results are in \cref{tab:result}. Our method has demonstrated significant improvements in both HumanArt and MSCOCO datasets. On HumanArt, our approach achieved a 2.26\% and 2.86\% AP and AP$\ddagger$ performance boost when compared to the current SOTA using the ViT-H backbone. This outcome serves as compelling evidence for the beneficial impact of the language model we have introduced in facilitating cross-domain knowledge integration. 

During testing on MSCOCO, our approach has the capability to revert to the original model weights by removing the auxiliary branch and additional image prompts. We conduct evaluations using these restored weights, so fine-tuning does not lead to a performance decrease on MSCOCO. This flexibility allows us to seamlessly switch between fine-tuned and original weights based on the specific requirements of our tasks without compromising performance on MSCOCO. As a result, on MSCOCO, our method demonstrates a substantial 3.74\% AP and 4.07\% AP$\ddagger$ improvement over the current SOTA. 

In order to verify the universality of our method to the model, we also conduct experiments based on swin-transformer~\cite{swin}, PVT~\cite{pvt} and PVTv2~\cite{pvtv2} backbone, with our corresponding model called SwinVLPose and PVLT. As shown in \cref{tab:result}, SwinVLPose-L and PVLT-B demonstrated improvements of 0.96\% and 1.67\% respectively on HumanArt. Results show that our approach demonstrates a consistent improvement trend, showcasing its general applicability across various model architectures.

Furthermore, we provide a performance analysis for each category. Our method shows a particularly significant improvement on ViT-H in challenging art categories, including shadow play (5.3\%), sketch (3.4\%), stained glass (4.7\%), and ukiyoe (8.2\%). The same trend was observed across different model sizes, indicating that our method demonstrates strong generalization capabilities for challenging artistic human poses. For a more detailed breakdown of performance by category, please refer to the appendix for detailed information.

In \cref{fig:vis}, we present a qualitative comparison between our method and the current SOTA. We provide 1-2 visualizations for each category to demonstrate the strong adaptability of our approach to different classes. The current SOTA still performs poorly in challenging tasks involving real human bodies with large motion amplitudes and artistic human bodies, such as human feet in \cref{fig:vis}\uppercase\expandafter{\romannumeral1}(a), shoulders in \cref{fig:vis}\uppercase\expandafter{\romannumeral1}(c), and eyes in \cref{fig:vis}\uppercase\expandafter{\romannumeral3}(e). In contrast, our method demonstrates significant improvements across various difficult scenarios.


\section{Conclusion}
In conclusion, our research presents a novel framework, VLPose, which effectively bridges the domain gap between human-centric natural and artificial scenarios. Our approach, characterized by the integration of domain-specific textual knowledge, enhances the adaptability and robustness of pose estimation models across diverse scenarios. This work holds promise for expanding the application of vision models in a wide range of real-world and artificial settings, including virtual reality and augmented reality, thus contributing to the evolving landscape of technology and its applications. Leveraging the synergy of language and vision features, we achieved a substantial improvement of 2.26\% and 3.74\% on the HumanArt and MSCOCO datasets compared to the current state-of-the-art models. 

\clearpage
{
\bibliographystyle{ieee_fullname}
\bibliography{egbib}
}

\end{document}


\appendix
\section{Appendix}
This appendix comprises an exploration of additional vision-language encoder structures (\ref{sec:decoder}), detailed customized prompts for each category (\ref{sec:prompt}), and in-depth performance analysis for each category (\ref{sec:result}).

\subsection{Vision-Language Decoder}
\label{sec:decoder}
Due to space constraints, we have only introduced three empirically validated vision-language decoder structures in the paper. In the appendix, we further delve into additional strategies for fusing multi-modal knowledge. 

\begin{figure*}[h]
\includegraphics[width=0.99\linewidth, trim=30 75 40 20, clip]{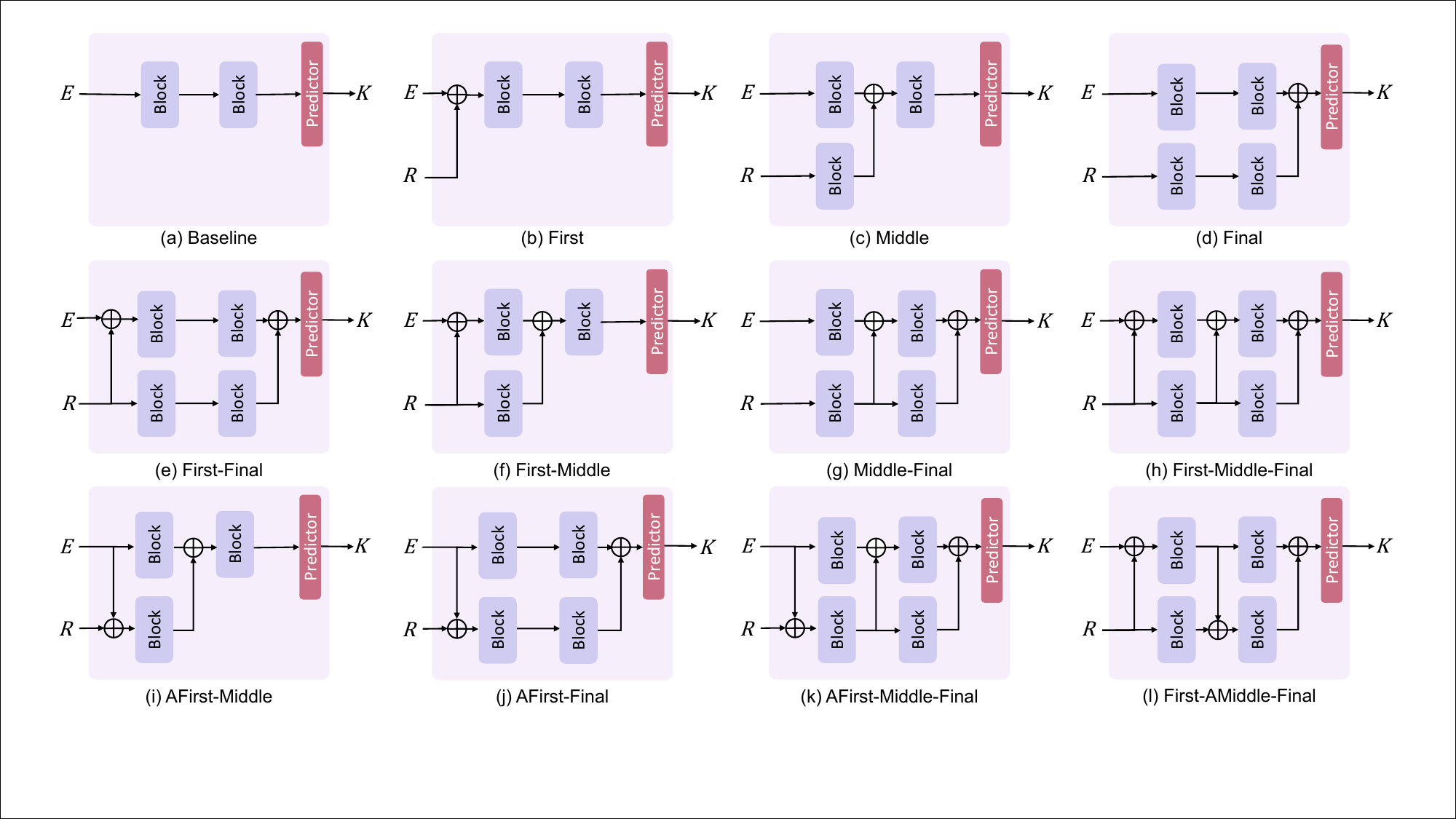}
\caption{A spectrum of decoder architectures. \textit{First}, \textit{Middle}, and \textit{Final} signify various positions of multimodal knowledge fusion. \textit{First} refers to the point before the first block, \textit{Middle} signifies the fusion between the two blocks, and \textit{Final} represents the stage after the second block. By default, knowledge is integrated from the auxiliary branch into the primary branch, denoted as \textit{A} for integration from the primary branch into the auxiliary branch.}
\label{fig:decoder}
\end{figure*}

The vision-language decoder comprises a primary branch and an auxiliary branch, each composed of two blocks. Each block includes a deconvolutional layer, batch normalization, and ReLU activation. The blocks perform upsampling and use a convolution layer with a $1\times1$ kernel to generate localization heatmaps for key points. We use the terms \textit{First}, \textit{Middle}, and \textit{Final} to represent different positions of multi-modal knowledge fusion, denoting before the first block, between the two blocks, and after the second block, respectively. By default, knowledge is injected from the auxiliary branch into the primary branch, denoted as \textit{A} for injecting from the primary branch into the auxiliary branch. Following this naming convention, we explore various vision-language decoder structures as illustrated in \cref{fig:decoder}.

We conducted ablation experiments for these structures, and the results are presented in \cref{tab:decoder}. The findings indicate that the \textit{First-Middle-Final-Injector} structure is the most beneficial for fusing knowledge from both modalities. Consequently, we adopt this structure, visually depicted in \cref{fig:decoder}(l). In the main text, this structure is referred to by the simplified name \textit{Dual Extractor-Injector Decoder}.

\begin{table}[h]
\centering
\setlength{\tabcolsep}{1mm}
\begin{tabular}{c|ccccc}
\toprule
model & AP & AP50 & AP75 & AR & AR50 \\
\midrule
Baseline & 36.45 & 52.60 & 38.56 & 43.12 & 59.28 \\
First & 36.93 & 52.86 & 39.34 & 43.48 & 59.55 \\
Final & 37.09 & \textbf{53.36} & 39.22 & 43.70 & \textbf{60.01} \\
Middle-Final & 37.01 & 52.81 & 39.28 & \uline{43.73} & 59.86 \\
First-Final & \uline{37.13} & 53.08 & \textbf{39.47} & 43.63 & 59.48 \\
First-Middle & 37.01 & \uline{53.27} & 39.19 & 43.57 & 59.53 \\
First-Middle-Final & 37.02 & 52.88 & 39.35 & 43.54 & 59.59 \\
AFirst-Middle-Final & 36.98 & 52.92 & 39.33 & 43.51 & 59.61 \\
AFirst-Final & 37.10 & 53.02 & 39.40 & 43.57 & 59.66 \\
AFirst-Middle & 37.02 & 53.21 & 39.28 & 43.62 & 59.89 \\
\rowcolor{gray!20}First-AMiddle-Final & \textbf{37.16} & 53.14 & \uline{39.43} & \textbf{43.76} & \uline{59.90} \\
\bottomrule
\end{tabular}
\caption{Ablation experiments on various vision-language pose decoder structure. \textit{First}, \textit{Middle}, and \textit{Final} signify various positions of multimodal knowledge fusion. \textit{First} refers to the point before the first block, \textit{Middle} signifies the fusion between the two blocks, and \textit{Final} represents the stage after the second block. By default, knowledge is integrated from the auxiliary branch into the primary branch, denoted as \textit{A} for integration from the primary branch into the auxiliary branch.}
\label{tab:decoder}
\end{table}

\subsection{Text prompt}
\label{sec:prompt}
In the main text, we have demonstrated the significant impact of tailoring prompts to individual semantic contexts, resulting in notable performance improvements. In this section, we provide an example list of customized prompts for each specific category, as outlined in \cref{tab:prompt}.

\begin{table}[h]
\centering
\setlength{\tabcolsep}{2mm}
\begin{tabular}{c|l|l}
\toprule
id & category & text prompt \\
\midrule
1 & \textit{cartoon} & \textit{ a cartoon human} \\
2 & \textit{digital art} & \textit{ a digital art human } \\
3 & \textit{ink painting} & \textit{ a ink-painting human} \\
4 & \textit{kids drawing} & \textit{ a kids-drawing human} \\
5 & \textit{mural} & \textit{ a mural human} \\
6 & \textit{oil painting} & \textit{ a oil-painting human} \\
7 & \textit{shadow play} & \textit{ a shadow-play human} \\
8 & \textit{sketch} & \textit{ a sketch human} \\
9 & \textit{stained glass} & \textit{ a stained glass human} \\
10 & \textit{ukiyoe} & \textit{ a ukiyoe human} \\
11 & \textit{watercolor} & \textit{ a watercolor human} \\
12 & \textit{garage kits} & \textit{ a garage-kits human} \\
13 & \textit{relief} & \textit{ a relief human} \\
14 & \textit{sculpture} & \textit{ a sculpture human} \\
15 & \textit{acrobatics} & \textit{ a acrobaticsing human photo} \\
16 & \textit{cosplay} & \textit{ a cosplaying human photo} \\
17 & \textit{dance} & \textit{ a dancing human photo} \\
18 & \textit{drama} & \textit{ a photo of a human in a drama} \\
19 & \textit{movie} & \textit{ a photo of a human in a movie} \\
\bottomrule
\end{tabular}
\caption{Individually tailored prompt examples for each distinct category.}
\label{tab:prompt}
\end{table}

\subsection{Detailed performance}
\label{sec:result}
We present our performance on each HumanArt category in \cref{tab:result} to provide a more detailed insight. The correspondence between category numbers and categories is shown in \cref{tab:prompt}. Our method shows a particularly significant improvement in ViT-H in challenging art categories, including shadow play (5.3\%), sketch (3.4\%), stained glass (4.7\%), and ukiyoe (8.2\%). The same trend was observed across different model sizes, indicating that our method demonstrates strong generalization capabilities for challenging artistic human poses.

\begin{table*}[tp]
\centering
\small
\setlength{\tabcolsep}{0.7mm}
\begin{tabular}{c|ccccccccccccccccccc|c}
\toprule
Method & 1 & 2 & 3 & 4 & 5 & 6 & 7 & 8 & 9 & 10 & 11 & 12 & 13 & 14 & 15 & 16 & 17 & 18 & 19 & Average \\
\midrule
ViTPose-S (sc) & 17.3 &  22.4 & 6.9 & 14.7 & 13.0 & 28.2 & 11.2 &  14.0 & 9.0 & 15.6 & 26.4 & 42.5 & 14.1 &  39.3 & 37.1 & 66.6 & 37.5 & 40.8 & 44.2 & 22.8 \\
ViTPose-S (ft) & 24.5 &  25.5 & 8.1 & 26.6 & 16.2 & 31.2 & 35.9 &  22.6 & 17.9 & 33.4 & 28.7 & 69.6 & 19.1 &  46.7 & 45.7 & 69.2 & 41.0 & 42.0 & 45.1 & 28.1 \\
ViTPose-S (pt) & 23.3 &  24.0 & 7.4 & 24.4 & 16.0 & 29.8 & 30.4 &  21.9 & 15.9 & 26.3 & 29.0 & 62.4 & 18.0 &  44.7 & 41.5 & 67.3 & 39.3 & 42.4 & 44.8 & 28.1 \\ 
\rowcolor{gray!20}VLPose-S & 26.2 &  24.3 & 7.9 & 26.9 & 17.0 & 31.4 & 33.5 &  27.5 & 19.4 & 34.1 & 30.6 & 68.3 & 21.2 &  46.0 & 43.4 & 69.4 & 40.9 & 42.0 & 44.8 & 30.6 \\
ViTPose-B (sc) & 20.5 &  25.5 & 7.6 & 22.0 & 16.4 & 31.8 & 12.4 &  17.5 & 12.9 & 22.4 & 30.2 & 52.3 & 17.6 &  44.6 & 44.1 & 72.0 & 42.6 & 44.2 & 49.2 & 27.0 \\
ViTPose-B (ft) & 28.6 &  28.3 & 8.9 & 32.9 & 19.8 & 34.9 & 38.5 &  27.5 & 21.9 & 38.5 & 33.4 & 74.5 & 22.9 &  50.1 & 51.4 & 74.8 & 45.5 & 46.3 & 48.8 & 32.6 \\
ViTPose-B (pt) & 21.9 &  23.6 & 6.3 & 19.8 & 14.5 & 28.4 & 31.7 &  15.3 & 14.9 & 24.1 & 26.6 & 63.9 & 14.8 &  42.7 & 40.9 & 67.2 & 37.8 & 40.6 & 43.5 & 26.9 \\
\rowcolor{gray!20}VLPose-B & 29.3 &  28.3 & 9.3 & 30.9 & 20.4 & 35.2 & 36.6 &  30.9 & 22.3 & 38.5 & 33.5 & 73.3 & 25.6 &  49.8 & 50.6 & 73.9 & 45.1 & 46.3 & 48.4 & 34.3 \\
ViTPose-L (sc) & 27.4 &  31.3 & 10.5 & 32.1 & 21.4 & 37.8 & 13.4 &  25.6 & 17.7 & 32.8 & 37.8 & 68.5 & 26.2 &  54.3 & 56.9 & 76.8 & 51.5 & 50.5 & 54.3 & 34.2 \\
ViTPose-L (ft) & 34.6 &  32.5 & 10.5 & 39.0 & 23.9 & 40.5 & 38.3 &  32.6 & 26.2 & 44.4 & 39.7 & 79.3 & 30.0 &  57.8 & 62.6 & 79.1 & 53.6 & 52.4 & 55.4 & 38.5 \\
ViTPose-L (pt) & 29.0 &  28.9 & 8.1 & 29.0 & 18.9 & 36.2 & 34.5 &  18.2 & 20.7 & 31.7 & 33.2 & 71.9 & 21.7 &  52.9 & 53.0 & 75.1 & 48.7 & 48.2 & 52.7 & 32.8 \\
\rowcolor{gray!20}VLPose-L & 36.1 &  33.2 & 11.2 & 35.9 & 24.5 & 41.0 & 40.7 &  37.3 & 27.3 & 47.3 & 40.0 & 79.3 & 32.4 &  58.0 & 61.9 & 78.9 & 53.0 & 50.9 & 54.2 & 40.2 \\
ViTPose-H (sc) & 31.8 &  33.6 & 11.7 & 36.4 & 23.8 & 40.6 & 16.8 &  32.0 & 21.6 & 37.2 & 40.2 & 72.8 & 30.7 &  55.4 & 60.7 & 80.1 & 54.8 & 52.4 & 55.2 & 37.7 \\
ViTPose-H (ft) & 37.2 &  35.2 & 11.2 & 40.3 & 25.4 & 42.3 & 37.1 &  36.1 & 26.6 & 46.0 & 41.4 & 80.0 & 32.8 &  58.9 & 64.8 & 81.0 & 56.5 & 55.0 & 57.9 & 40.8 \\
ViTPose-H (pt) & 36.5 &  34.8 & 11.2 & 39.2 & 25.5 & 42.1 & 35.3 &  35.7 & 25.1 & 43.0 & 41.4 & 78.4 & 33.5 &  59.2 & 63.2 & 80.1 & 56.0 & 55.4 & 58.2 & 40.8 \\
\rowcolor{gray!20}VLPose-H & 38.4 &  36.0 & 12.2 & 39.3 & 26.7 & 44.4 & 40.6 &  39.1 & 29.8 & 51.2 & 42.6 & 81.2 & 35.2 &  61.5 & 64.5 & 81.6 & 56.9 & 54.7 & 58.1 & 43.0 \\
\bottomrule
\end{tabular}
\caption{Comparison of Average Precision (AP) of each category among widely used human pose estimation models. The correspondence between category numbers and categories is shown in \cref{tab:prompt}. \textit{sc}: scratch model pre-trained on COCO. \textit{ft}: finetune the last layer of the model on HumanArt. \textit{ft}: prompt tuning on HumanArt.}
\label{tab:result}
\end{table*}


